\pdfoutput=1

\documentclass[11pt]{article}

\usepackage{acl}
\usepackage{amsmath}
\usepackage{bm}
\usepackage{amsfonts}
\usepackage{booktabs}
\usepackage{multirow}
\usepackage{times}
\usepackage{latexsym}
\usepackage{cleveref}
\usepackage{subfigure}

\usepackage[T1]{fontenc}

\usepackage[utf8]{inputenc}

\usepackage{microtype}

\title{A Simple and Plug-and-play Method for Unsupervised Sentence Representation Enhancement}

\author{Lingfeng Shen \\ Johns Hopkins Iniversity
        \And  Haiyun Jiang \\ Tencent AI Lab \And Lemao Liu \\ Tencent AI Lab \And
        Shuming Shi \\ Tencent AI Lab}

\begin{document}
\maketitle
\begin{abstract}
Generating proper embedding of sentences through an
unsupervised way is beneficial to semantic matching and retrieval problems in real-world scenarios.
This paper presents \textbf{Rep}resentation \textbf{AL}chemy (RepAL), an extremely simple post-processing method that enhances sentence representations.
The basic idea in RepAL is to de-emphasize redundant information of sentence embedding generated by pre-trained models. 
Through comprehensive experiments, we show that RepAL is free of training and is a plug-and-play method that can be combined with most existing unsupervised sentence learning models. 
We also conducted in-depth analysis to understand RepAL.






\end{abstract}

\section{Introduction}
Learning high-quality sentence embeddings is a fundamental task in Natural Language Processing (NLP) field \cite{socher2011dynamic,le2014distributed,kiros2015skip,reimers2019sentence,gao2021simcse}. 
In real-world scenarios, especially when a large amount of supervised data is unavailable, an approach that provides high-quality sentence embeddings in an unsupervised paradigm is of great value.

Generally, the unsupervised sentence encoder (USE) can be categorized into two paradigms. 
The first is pre-trained language model (PTM) \cite{devlin2019bert,liu2019roberta} based paradigm, which are naturally good unsupervised sentence representation learning models.
For example, BERT \cite{devlin2019bert} and BERT-like \cite{liu2019roberta,he2020deberta,raffel2020exploring} models, commit to design stronger pre-trained language models by self-training with mask or next sentence prediction. 
While designing stronger PTMs for better sentence representation is extremely expensive, time-consuming, and labor-intensive.
Based on PTMs, secondary trained, e.g., contrastive-based methods \cite{reimers2019sentence,logeswaran2018efficient,gao2021simcse}, proved to be effective to further improve the representation quality of sentences.
For example, SimCSE \cite{gao2021simcse} minimizes the distance between positive pairs of sentences and pulls away from the negative pairs of sentences in the embedding space, which achieves promising performance. 
This paper focuses on enhancing sentence embeddings generated by above two paradigms in an supervised way. 
Our basic idea is to \emph{refine sentence representations by removing the redundant information from the levels of sentences and corpus}. Considering corpus-level, there may be shared information among the corpus, which may lead to homogenous properties for all sentence embeddings, which diminishes the distinctiveness between sentences.
As for sentence-level, there are several trivial words within the sentence. 
Such words are proven to bring a negative impact on downstream NLP tasks, like Natural Language Inference (NLI) \cite{mahabadi2020end,zhou2020towards} and text classification \cite{choi2020less,qian2021counterfactual}. 

Therefore, we propose a simple, straightforward, and effective method called representation alchemy (RepAL), which improves sentence representations without training and extra resources. 
RepAL accepts raw sentence representations as inputs, which are generated from existing unsupervised sentence models.
Then RepAL outputs refined representations by extracting two redundant representations from different perspectives.
Intuitively, it is like \emph{an alchemy that improves sentence representation by refinement}.
It’s worth mentioning that our proposed RepAL can be applied to almost USEs and is a plug-and-play method in sentence embedding enhancement without extra training cost. 
To verify, we perform extensive experiments on both English and Chinese benchmarks, and results demonstrate the effectiveness of the proposed RepAL.




\section{Related Work}
Methods for unsupervised sentence learning have been extensively explored. 
Early works are mainly based on distributional hypothesis \cite{socher2011dynamic,le2014distributed}. 
Hill \cite{hill2016learning} proposed to learn sentence representations with its internal structure. 
Then Pagliardini \cite{pagliardini2018unsupervised} proposed Sent2Vec, a simple unsupervised model allowing to compose sentence embeddings using word vectors.

Then strong pre-trained language model \cite{devlin2019bert} emerged from the blue. Such pre-trained models own potential to improve the quality of sentence representation. However, models like BERT own strong anisotropy in their embedding space, which means the sentence embeddings produced by BERT have extremely high cosine similarity, leading to unsatisfactory performances on sentence embedding.

Recently, contrastive learning began to play an important role in unsupervised sentence representation learning \cite{zhang2020unsupervised,yan2021consert,meng2021coco,gao2021simcse,wang2021cline}. Such methods are based on the assumption that high-quality embedding methods should bring similar sentences closer while pushing away dissimilar ones. 

Specifically, the most relevant work is BERT-whitening \cite{su2021whitening,su2021whitening}, a post-processing method, and a detailed comparison between it and our work is illustrated in Appendix~\ref{diff}.


\section{Methodology}
\subsection{Problem Formulation}
In unsupervised sentence representation learning, we take a collection of unlabeled sentences $\{x_{i}\}_{i=1}^{n}$, also we choose a suitable unsupervised sentence learning model (e.g., BERT) as the encoder $f(\cdot;\theta)$, where $\theta$ represents the trainable parameters in $f$. 
Specifically, we have a carefully designed training objective $\mathcal{L}$ for unsupervised training, and $\theta$ is then fixed as $\theta_{0}$ where $\theta_{0} = \operatorname{argmin}\mathcal{L}$. 
Finally, we obtain the sentence representation $v_{i}$ for $x_i$ by feeding it into the encoder, i.e.,  $v_{i} = f(x_{i};\theta_{0})$.

RepAL plays its role in refining $v_i$ to $v_i'$ with $v_i' = g(v_i)$, instead of directly selecting $v_{i}$ for sentence representation.
RepAL aims to extract and refine two types of redundancy, namely \emph{sentence-level redundancy} and \emph{corpus-level redundancy}, respectively.
Sentence-level redundancy denotes the useless word information hidden in the target sentence, which may bias the representations that reflect the core semantics of the sentence.
Corpus-level redundancy denotes the shared redundant information in all sentence representations within the dataset, making all the representations tend to be homogenous and thus reducing the distinction.

RepAL generates $x_{i}^{*}$ by an operation called \emph{partial mask} on $x_{i}$, then feed $x_{i}^{*}$ into the encoder $f(.;\theta_{0})$ to obtain sentence-level redundancy embeddings $v_{i}^{*}$. 
Besides, RepAL produces a global vector $\hat{v}$ as corpus-level redundancy embedding.
Finally, RepAL generates the refined embedding $v_i'$ for downstream tasks through the \emph{embedding refinement} operation by combining $v_{i}, v_{i}^{*}$ and $\hat{v}$. 


\subsection{Redundant Embedding Generation}
In RepAL, we firstly detect redundant information and generate their embeddings from the target sentence, which is a groundbreaking step in our method and determines the performance. 
Moreover, we also conduct deep analyses towards two kinds of redundancy, and defer them to Appendix~\ref{analysis}.

\subsubsection{Sentence-level Redundancy}
We apply a partial mask to extract the sentence-level redundancy. 
Specifically, given a sentence $x_{i}=\{w_{1},w_{2},\dots,w_{N}\}$, partial mask generates a partially masked sentence $x_{i}^{*}$, a mask version of $x_{i}$, where \emph {informative} words in $x_{i}$ are replaced with [MASK] to distill the trivial words from the sentence. 
We judge the words as informative according to their TF-IDF \cite{luhn1958automatic,jones1972statistical} values\footnote{In RepAL, we use the default function (based on TF-IDF) in Jieba toolkit to extract the keywords within a sentence.} calculated on a general corpus. 
In the following, we generate partially masked sentence $x_{i}^{*}$, where only the keywords in the sentence $x_{i}$ are masked, and $f(x_{i}^{*})$ as the corresponding redundant embedding.
Since the model is forced to see only the non-masked context words, $f(x_{i}^{*})$ actually encode the information from the trivial words (non-keywords), and the sentence-level redundant embedding is defined as follows:
\begin{equation}
    x_{i}^{*}=\text{PartialMask}(x_{i},\text{keyword}); \quad
    v_{i}^{*}=f(x_{i}^{*})
\end{equation}
where $v_{i}^{*}$ is the sentence-level redundant embedding of $x_{i}$. Diminishing $v_{i}^{*}$ from $x_{i}$ representation aims to de-emphasizes the influence of trivial words when producing sentence embedding.

 
\subsubsection{Corpus-level Redundancy}
Given a sentence set $\mathcal{X}=\{x_{i}\}_{i=1}^{n}$, we feed all the sentences to the encoder $f$, and take the average embedding $\hat{v}$ as its corpus-level redundant embedding, which can be formally defined as follows:
\begin{equation}
    \hat{v}= \frac{\Sigma_{i=1}^{n} f(x_{i})}{n} 
\end{equation}
where $\hat{v}$ is the corpus-level redundant embedding. Diminishing $\hat{v}$ from each sentence's representation makes them more distinguishable to each other.

\subsection{Embedding Refinement}
After generating redundant embedding, the embedding refinement operation can be formalized via the conceptually simple subtraction operation, which is defined as follows:
\begin{equation}
   v_i' = f(x_{i}) - \lambda_{1} \cdot v_{i}^{*} - \lambda_{2} \cdot \hat{v}
\end{equation}
where $f(x)$ corresponds to the original embedding of $x_{i}$, and $v_{i}^{*}$ and $\hat{v}$ represent the redundant embedding at two levels, respectively. Since the two redundant embeddings typically do not contribute completely equal to the embedding $v$, we introduce two independent hyper-parameters $\lambda_{1}$ and $\lambda_{2}$ to balance the two terms.

\begin{table*}[!h]\footnotesize
\centering
\begin{tabular}{@{}c|ccccc|c@{}}
\toprule
Baseline     & ATEC  & BQ    & LCQMC & PAWSX & STS-B & Avg   \\ \midrule
BERT& 16.51$\rightarrow$\color[HTML]{3166FF}19.58  & 29.35$\rightarrow$\color[HTML]{3166FF}32.89 & 41.71$\rightarrow$\color[HTML]{3166FF}44.53 & 9.84$\rightarrow$\color[HTML]{3166FF}11.28  & 34.65$\rightarrow$\color[HTML]{3166FF}47.00 & 26.41$\rightarrow$\color[HTML]{3166FF}31.06(+4.65) \\
RoBERTa & 24.61$\rightarrow$\color[HTML]{3166FF}27.00 & 40.54$\rightarrow$39.51 & 70.55$\rightarrow$\color[HTML]{3166FF}70.98  & 16.23$\rightarrow$\color[HTML]{3166FF}16.98 & 63.55$\rightarrow$\color[HTML]{3166FF}64.01 & 43.10$\rightarrow$\color[HTML]{3166FF}43.70(+0.60) \\ 
RoFormer     & 24.29$\rightarrow$\color[HTML]{3166FF}25.07 & 41.91$\rightarrow$\color[HTML]{3166FF}42.56 & 64.87$\rightarrow$\color[HTML]{3166FF}65.33 & 20.15$\rightarrow$20.13 & 56.65$\rightarrow$\color[HTML]{3166FF}57.23 & 41.57$\rightarrow$\color[HTML]{3166FF}42.06(+0.49)  \\ 
NEZHA        & 17.39$\rightarrow$\color[HTML]{3166FF}18.98 & 29.63$\rightarrow$\color[HTML]{3166FF}30.53 & 40.60$\rightarrow$\color[HTML]{3166FF}41.85 & 14.90$\rightarrow$\color[HTML]{3166FF}15.43 & 35.84$\rightarrow$\color[HTML]{3166FF}36.68 & 27.67$\rightarrow$\color[HTML]{3166FF}28.69(+1.02) \\ \midrule
W-BERT      & 20.61$\rightarrow$\color[HTML]{3166FF}23.29 & 25.76$\rightarrow$\color[HTML]{3166FF}29.83 & 48.91$\rightarrow$\color[HTML]{3166FF}50.01 & 16.82$\rightarrow$\color[HTML]{3166FF}16.96 & 61.19$\rightarrow$\color[HTML]{3166FF}61.46 & 34.66$\rightarrow$\color[HTML]{3166FF}36.31(+1.65)  \\ 
W-RoBERTa      & 29.59$\rightarrow$\color[HTML]{3166FF}30.44 & 28.95$\rightarrow$\color[HTML]{3166FF}43.12 & 70.82$\rightarrow$\color[HTML]{3166FF}71.39 & 17.99$\rightarrow$\color[HTML]{3166FF}18.48 & 69.19$\rightarrow$\color[HTML]{3166FF}70.92 & 43.31$\rightarrow$\color[HTML]{3166FF}46.87(+2.56)  \\ 
W-RoFormer      & 26.04$\rightarrow$\color[HTML]{3166FF}27.68 & 28.13$\rightarrow$\color[HTML]{3166FF}42.63 & 60.92$\rightarrow$\color[HTML]{3166FF}61.55 & 23.08$\rightarrow$23.05 & 66.96$\rightarrow$\color[HTML]{3166FF}67.13 & 41.03$\rightarrow$\color[HTML]{3166FF}44.38(+3.35)  \\ 
W-NEZHA      & 18.83$\rightarrow$\color[HTML]{3166FF}21.33 & 21.94$\rightarrow$\color[HTML]{3166FF}23.02 & 50.52$\rightarrow$\color[HTML]{3166FF}52.01 & 18.15$\rightarrow$\color[HTML]{3166FF}19.00 &60.84$\rightarrow$60.82 & 34.06$\rightarrow$\color[HTML]{3166FF}35.24(+1.18)  \\\midrule
C-BERT      & 26.35$\rightarrow$\color[HTML]{3166FF}28.69 & 46.68$\rightarrow$\color[HTML]{3166FF}48.02 & 69.22$\rightarrow$\color[HTML]{3166FF}69.98 & 10.89$\rightarrow$\color[HTML]{3166FF}12.03 & 68.89$\rightarrow$\color[HTML]{3166FF}69.66 & 44.41$\rightarrow$\color[HTML]{3166FF}45.68(+1.27)  \\ 
C-RoBERTa      & 27.39$\rightarrow$\color[HTML]{3166FF}28.43 & 47.20$\rightarrow$\color[HTML]{3166FF}47.14 & 67.34$\rightarrow$\color[HTML]{3166FF}67.98 & 09.36$\rightarrow$\color[HTML]{3166FF}10.55 & 72.02$\rightarrow$71.80 & 44.66$\rightarrow$\color[HTML]{3166FF}45.18(+0.52)  \\ 
C-RoFormer      & 26.24$\rightarrow$\color[HTML]{3166FF}27.68 & 47.13$\rightarrow$\color[HTML]{3166FF}47.63 & 66.92$\rightarrow$\color[HTML]{3166FF}67.85 & 11.08$\rightarrow$\color[HTML]{3166FF}11.65 & 69.84$\rightarrow$\color[HTML]{3166FF}69.73 & 44.24$\rightarrow$\color[HTML]{3166FF}44.91(+0.67)  \\ 
C-NEZHA      & 26.02$\rightarrow$\color[HTML]{3166FF}26.73 & 47.44$\rightarrow$\color[HTML]{3166FF}48.02 & 70.02$\rightarrow$\color[HTML]{3166FF}70.63 & 11.46$\rightarrow$\color[HTML]{3166FF}11.80 &68.97$\rightarrow$\color[HTML]{3166FF}69.53 & 44.78$\rightarrow$\color[HTML]{3166FF}45.34(+0.56)  \\ \midrule
Sim-BERT      & 33.14$\rightarrow$\color[HTML]{3166FF}33.48 & 50.67$\rightarrow$\color[HTML]{3166FF}51.14 & 69.99$\rightarrow$\color[HTML]{3166FF}72.44 & 12.95$\rightarrow$\color[HTML]{3166FF}13.58 & 69.04$\rightarrow$\color[HTML]{3166FF}69.55 & 47.16$\rightarrow$\color[HTML]{3166FF}48.04(+0.88)  \\ 
Sim-RoBERTa      & 32.23$\rightarrow$\color[HTML]{3166FF}33.10 & 50.61$\rightarrow$\color[HTML]{3166FF}51.53 & 74.22$\rightarrow$\color[HTML]{3166FF}74.77 & 12.25$\rightarrow$\color[HTML]{3166FF}13.28 & 71.13$\rightarrow$\color[HTML]{3166FF}72.20 & 48.09$\rightarrow$\color[HTML]{3166FF}48.98(+0.89)  \\ 
Sim-RoFormer      & 32.33$\rightarrow$\color[HTML]{3166FF}32.59 & 49.13$\rightarrow$\color[HTML]{3166FF}49.46 & 71.61$\rightarrow$\color[HTML]{3166FF}72.13 & 15.25$\rightarrow$\color[HTML]{3166FF}15.69 & 69.45$\rightarrow$\color[HTML]{3166FF}70.01 & 47.55$\rightarrow$\color[HTML]{3166FF}48.02(+0.47)  \\ 
Sim-NEZHA      & 32.14$\rightarrow$\color[HTML]{3166FF}32.52 & 46.08$\rightarrow$\color[HTML]{3166FF}47.42 & 60.38$\rightarrow$\color[HTML]{3166FF}60.51 & 16.60$\rightarrow$\color[HTML]{3166FF}16.58 &68.50$\rightarrow$\color[HTML]{3166FF}69.19 & 44.74$\rightarrow$\color[HTML]{3166FF}45.26(+0.52)  \\
\bottomrule
\end{tabular}
\caption{The experimental results of RepAL on Chinese semantic similarity benchmarks. The numbers before $\rightarrow$ indicate the performance without RepAL and the numbers after $\rightarrow$ mean the performance with RepAL. Blue numbers indicate RepAL improves the baseline.}
\label{table:1}
\end{table*}

\section{Experiments}
This section shows that our method can be adaptive to various USE and improves their performance.

\subsection{Baselines and Benchmarks}
To verify the effectiveness of our method, we evaluate RepAL on both Chinese and English benchmarks. 
To investigate whether our method can be applied to various unsupervised sentence encoder (USE), we choose two kinds of encoders: vanilla USE and secondary trained USE. 
For vanilla USE, we select BERT \cite{devlin2019bert}, RoBerTa \cite{liu2019roberta}, RoFormer \cite{su2021roformer} and NEZHA \cite{wei2019nezha} for Chinese; for English, we select BERT$_{base}$, BERT$_{large}$ \cite{devlin2019bert} and RoBERTa$_{base}$ \cite{reimers2019sentence}. Specifically, we name the secondary trained USE equipped with whitening \cite{huang2021whiteningbert}, ConSERT \cite{yan2021consert}, and SimCSE \cite{gao2021simcse} as W-USE (e.g., W-BERT), C-USE (e.g., C-BERT), and Sim-USE (e.g., Sim-BERT), respectively. Results of Sim-USE and C-USE are from our implementation. Details of training SimCSE on Chinese benchmarks are deferred to Appendix~\ref{cn}.
\begin{itemize}
    \item \textbf{Chinese}: We select five Chinese benchmarks for evaluation: AETC, LCQMC, BQ, PAWSX, and STS-B. The details about them are deferred to the Appendix~\ref{cn}.
    \item \textbf{English}: We select STS task benchmarks as our English datasets, including STS 2012-2016 tasks \cite{agirre2012semeval,agirre2013sem,agirre2014semeval,agirre2015semeval,agirre2016semeval}, the STS benchmark \cite{cer2017semeval} and the SICK-Relatedness dataset \cite{marelli2014sick}. 
\end{itemize}
\subsection{Experimental Settings}
The vanilla USE in our experiments is the same as their original settings. 
Besides, we keep the settings baselines the same as their original ones. 
As for hyper-parameters, we follow previous unsupervised work \cite{gao2021simcse,yan2021consert}, and tune $\lambda_{1}$ and $\lambda_{2}$ based on the STS-B dev set. 
The results are evaluated through Spearman correlation.

\begin{table*}[!h]\small
\centering
\begin{tabular}{@{}c|ccccc|c@{}}
\toprule
Baseline     & STS-12  & STS-13    & STS-14 & STS-15 &STS-16 & Avg   \\ \midrule
BERT& 57.86$\rightarrow$\color[HTML]{3166FF}59.55 & 61.97$\rightarrow$\color[HTML]{3166FF}66.20 & 62.49$\rightarrow$\color[HTML]{3166FF}65.19 & 70.96$\rightarrow$\color[HTML]{3166FF}73.50  & 69.76$\rightarrow$\color[HTML]{3166FF}72.10 & 63.69$\rightarrow$\color[HTML]{3166FF}66.70(+3.01) \\
BERT$_{l}$ & 57.74$\rightarrow$\color[HTML]{3166FF}59.90 & 61.16$\rightarrow$\color[HTML]{3166FF}66.20 & 61.18$\rightarrow$\color[HTML]{3166FF}65.62  & 68.06$\rightarrow$\color[HTML]{3166FF}73.01 & 70.30$\rightarrow$\color[HTML]{3166FF}74.72 & 62.62$\rightarrow$\color[HTML]{3166FF}67.47(+4.85) \\ 
RoBERTa     & 58.52$\rightarrow$\color[HTML]{3166FF}60.88 & 56.21$\rightarrow$\color[HTML]{3166FF}62.20 & 60.12$\rightarrow$\color[HTML]{3166FF}64.10 & 69.12$\rightarrow$\color[HTML]{3166FF}71.41 & 63.69$\rightarrow$\color[HTML]{3166FF}69.94 & 60.59$\rightarrow$\color[HTML]{3166FF}65.41 (+4.82)  \\ \midrule
W-BERT      & 63.62$\rightarrow$\color[HTML]{3166FF}64.50 & 73.02$\rightarrow$\color[HTML]{3166FF}73.69 & 69.23$\rightarrow$\color[HTML]{3166FF}69.69 & 74.52$\rightarrow$\color[HTML]{3166FF}74.69 &  72.15$\rightarrow$\color[HTML]{3166FF}76.11 & 69.21$\rightarrow$\color[HTML]{3166FF}70.39 (+1.18)  \\ 

W-BERT$_{l}$      & 63.62$\rightarrow$\color[HTML]{3166FF}63.90 &73.02$\rightarrow$\color[HTML]{3166FF}73.41 & 69.23$\rightarrow$\color[HTML]{3166FF}70.01 & 74.52$\rightarrow$\color[HTML]{3166FF}75.18 & 72.15$\rightarrow$\color[HTML]{3166FF}75.89 & 69.21$\rightarrow$\color[HTML]{3166FF}70.39 (+1.18)  \\ 

W-RoBERTa      & 68.18$\rightarrow$\color[HTML]{3166FF}68.85 & 62.21$\rightarrow$\color[HTML]{3166FF}63.03 & 67.13$\rightarrow$\color[HTML]{3166FF}67.69 & 67.63$\rightarrow$\color[HTML]{3166FF}68.23 & 74.78$\rightarrow$\color[HTML]{3166FF}75.44 & 67.17$\rightarrow$\color[HTML]{3166FF}68.43 (+1.26)  \\ \midrule

C-BERT      & 64.09$\rightarrow$\color[HTML]{3166FF}65.01 & 78.21$\rightarrow$\color[HTML]{3166FF}78.54 & 68.68$\rightarrow$\color[HTML]{3166FF}69.04 & 79.56$\rightarrow$\color[HTML]{3166FF}79.90 & 75.41$\rightarrow$\color[HTML]{3166FF}75.74 & 72.27$\rightarrow$\color[HTML]{3166FF}72.69 (+0.42)  \\ 
C-BERT$_{l}$      & 70.23$\rightarrow$\color[HTML]{3166FF}70.70 & 82.13$\rightarrow$\color[HTML]{3166FF}82.54 & 73.60$\rightarrow$\color[HTML]{3166FF}74.12 & 81.72$\rightarrow$\color[HTML]{3166FF}82.01 & 77.01$\rightarrow$\color[HTML]{3166FF}77.58 & 76.03$\rightarrow$\color[HTML]{3166FF}76.48 (+0.45)  \\ \midrule
Sim-BERT      & 68.93$\rightarrow$\color[HTML]{3166FF}69.33 & 78.68$\rightarrow$\color[HTML]{3166FF}78.93 & 73.57$\rightarrow$\color[HTML]{3166FF}73.95 & 79.68$\rightarrow$\color[HTML]{3166FF}80.01 & 79.11$\rightarrow$\color[HTML]{3166FF}79.29 & 75.11$\rightarrow$\color[HTML]{3166FF}75.44 (+0.33)  \\ 
Sim-BERT$_{l}$      & 69.25$\rightarrow$\color[HTML]{3166FF}69.60 & 78.96$\rightarrow$\color[HTML]{3166FF}79.30 & 73.64$\rightarrow$\color[HTML]{3166FF}73.92 & 80.06$\rightarrow$\color[HTML]{3166FF}80.31 & 79.08$\rightarrow$\color[HTML]{3166FF}79.42 & 75.31$\rightarrow$\color[HTML]{3166FF}75.61 (+0.30)  \\ 
\bottomrule
\end{tabular}
\caption{The experimental results of RepAL on English semantic similarity benchmarks. `Avg' indicates the average performance of all English benchmarks including STS-B and SICK-R in Table~\ref{table:21}, and BERT$_{l}$ means BERT$_{large}$ during the experiments.}
\label{table:2}
\end{table*}

\begin{table}[!h]\small
\centering
\begin{tabular}{@{}c|cc@{}}
\toprule
Baseline     & STS-B  & SICK-R      \\ \midrule
BERT& 59.04$\rightarrow$ \color[HTML]{3166FF}66.35& 63.75$\rightarrow$ \color[HTML]{3166FF}64.55  \\
BERT$_{l}$ & 59.59$\rightarrow$ \color[HTML]{3166FF}68.21& 60.34$\rightarrow$ \color[HTML]{3166FF}64.61  \\ 
RoBERTa     & 55.16$\rightarrow$ \color[HTML]{3166FF}65.75& 61.33$\rightarrow$ \color[HTML]{3166FF}63.61   \\ \midrule
W-BERT      & 71.34$\rightarrow$ \color[HTML]{3166FF}71.45& 60.60$\rightarrow$ \color[HTML]{3166FF}62.61  \\ 

W-BERT$_{l}$      & 71.34$\rightarrow$ \color[HTML]{3166FF}69.56& 60.60$\rightarrow$ \color[HTML]{3166FF}65.00  \\ 

W-RoBERTa      & 71.43$\rightarrow$ \color[HTML]{3166FF}72.03& 58.80$\rightarrow$ \color[HTML]{3166FF}63.95 \\ \midrule

C-BERT      & 73.12$\rightarrow$ \color[HTML]{3166FF}73.45& 66.79$\rightarrow$ \color[HTML]{3166FF}67.15  \\ 
C-BERT$_{l}$      & 77.48$\rightarrow$ \color[HTML]{3166FF}77.91& 70.02$\rightarrow$ \color[HTML]{3166FF}70.51  \\ \midrule
Sim-BERT      & 75.71$\rightarrow$ \color[HTML]{3166FF}76.00& 70.12$\rightarrow$ \color[HTML]{3166FF}70.51  \\ 
Sim-BERT$_{l}$      & 75.84$\rightarrow$ \color[HTML]{3166FF}76.11& 70.34$\rightarrow$ \color[HTML]{3166FF}70.61  \\ 
\bottomrule
\end{tabular}
\caption{The results of RepAL on STS-B and SICK-R}
\label{table:21}
\end{table}

\subsection{Results}
As shown in Table~\ref{table:1}, RepAL improves the baselines' performance in most cases. 
For example, RepAL produces 4.65\%, 1.65\%, 1.27\%, and 0.88\% improvement to BERT, W-BERT, C-BERT, and Sim-BERT, respectively. 
Generally, as USE becomes stronger, the improvements brought by RepAL decrease. 
Still, for strong baselines like C-BERT and Sim-BERT, RepAL still makes progress over them. Specifically, RepAL achieves $1.27\%$ and $0.88\%$ performance increase for C-BERT and Sim-BERT, indicating the effectiveness of RepAL on extremely strong baselines. 
The results on English benchmarks are listed in Table~\ref{table:2},~\ref{table:21}, where RepAL also obtains improvements over various USEs. Overall, both results on Chinese and English benchmarks demonstrate the effectiveness of RepAL and illustrate that RepAL is a plug-and-play method for sentence representation enhancement.

\subsection{Ablation study}
This section investigates the individual effective of embedding refinement of two different levels. As shown in Table~\ref{table:00}, each of operation is beneficial and combining them lead to stronger performance.

\begin{table}[!h]\small
\centering
\begin{tabular}{@{}c|cccc@{}}
\toprule
     & BERT  & W-BERT & C-BERT & S-BERT     \\ \midrule
RepAL&31.06&36.31&45.68&48.04\\
w/o Sen&29.28&34.03&43.49&46.09\\
w/o Cor&30.42&35.78&45.03&47.63\\
\bottomrule
\end{tabular}
\caption{Ablation studies of RepAL on Chinese benchmarks. `S-BERT' refers to `Sim-BERT'.}
\label{table:00}
\end{table}

\section{Conclusion}
In this paper, we propose RepAL, a universal method for unsupervised sentence representation enhancement. Based on the idea that de-emphasize redundant information, RepAL extracts then refines redundant information for the sentence embedding at sentence-level and corpus-level. Through a simple embedding refinement operation, RepAL successfully achieves improvements on both Chinese and English benchmarks and is proved to be a simple and plug-and-play method in modern techniques for unsupervised sentence representation.

\section{Limitation}
RepAL is a universal method for sentence representation, and sentence representation is eventually used for downstream tasks. However, RepAL does not consider task-specific information since different downstream tasks may have different preferences. Therefore, exploring task-specific modification on RepAL is a future direction.

\bibliography{acl_latex}
\bibliographystyle{acl_natbib}

\clearpage
\appendix

\section{Difference between BERT whitening}\label{diff}
Some post-processing methods have been proposed to improve the quality of contextual sentence embeddings to solve such a problem. The post-processing paradigm aims to enhance sentence embeddings through simple and efficient operations without extra training or data. The most promising method is whitening \cite{huang2021whiteningbert}, dedicated to transforming sentence embedding into Gaussian-like embedding, which proved to be effective in sentence embedding improvement.
Specifically, the most relevant work to ours is whitening \cite{huang2021whiteningbert} since the corpus-level refinement is similar to the average embedding subtraction in whitening. However, there are three principal differences between such two works. Firstly, the motivation is different. Whitening aims at transforming the sentence embedding to Gaussian-like embedding for distance measurement on an orthogonal basis. Our method starts a perspective of redundancy refinement, which aims to diminish the impact of trivial words within a sentence during similarity calculation. Second, the methodology is different. Our method additionally employs a partial mask to filter the redundancy and introduce weight factors to control the impact during embedding refinement. Lastly, the in-depth analysis shows that our method aims to diminish the upper bound of the largest eigenvalue of the embedding matrix and the impact of trivial words, which is irrelevant to whitening's effects.

\section{Details of Chinese benchmarks and training Chinese SimCSE}\label{cn}
(1) ATEC: A semantic similarity dataset related to customer service; (2) LCQMC: A dataset consisting problem matching across multiple domains; (3) BQ: a dataset consisting problem matching related to bank and finance; (4) PAWSX \cite{yang2019paws} : The dataset contains multilingual paraphrase and non-paraphrase pairs, we select the Chinese part; (5) STS-B: A Chinese benchmark labeled by semantic correlation between two sentences.

As for training SimCSE on benchmarks, we delete the labels in ATEC, LCQMC, BQ, PAWSX, and STS-B, and merge them into an unsupervised corpus. Then SimCSE is launched on the merged corpus, the best checkpoint is selected according to the STS-B dev set, which keeps the same setting in previous works \cite{gao2021simcse,yan2021consert}. Specifically, we find that the best dropout rate $r$ on Chinese corpus is about 0.3.

\section{Detailed Analysis and Discussion}\label{analysis}
The proposed RepAL enhances sentence embedding by filtering redundant information from two levels: sentence-level and corpus-level. 
Despite the presentations of the overall experiment results and analysis, the intrinsic properties of RepAL remain unclear. 
In this section, we illustrate the reasons why RepAL is effective in enhancing sentence embedding.

In Sec~\ref{wordlevel}, we provide the evidence about the impact of trivial words in sentence embedding and show the capacity of our sentence-level embedding refinement. 
In Sec~\ref{corpuslevel}, we show why the corpus-level embedding refinement enhances sentence embedding and illustrate the relation between the largest eigenvalue and performance. 

\subsection{Sentence-level Refinement}\label{wordlevel}
We conduct analyses for sentence-level refinement (SR) as follows: we investigate the impact of trivial words w/o RepAL, which explains the necessity of removing such redundancy information and validates the effectiveness of SR.


We first define the importance $H$ of word $w \in x_{i}$ in semantic similarity calculation, which can be defined as follows:
\begin{equation}\label{eq:h}
    H(x_{i}, x^{-}_{i}; w) = Sim(x_{i},x^{-}_{i})-Sim(x_{i}/w,x^{-}_{i})
\end{equation}
where $x_{i}$ and $x^{-}_{i}$ are a pair of sentences and $x_{i}/w_{i}$ means deleting the word $w_{i}$ from $x_{i}$. Note that we do not consider the words in $x^{-}_{i}$ since it is equivalent to evaluation on more sentences. Then we define the set of trivial words within $x_{i}$ as $S(x_{i})$, which are unmasked by jieba. Thus we can define the \textbf{redundancy overlap ratio} $r(p_{i})$ of a sentence pair $p_{i} = (x_{i},x^{-}_{i})$ as follows:
\begin{equation}
    r(p_{i}) = \frac{|S(x_{i})\cap T(x_{i})|}{|T(x_{i})|}
\end{equation}
where $T(x_{i})$ represents the top-5 words with highest importance $H$ in $x_{i}$. $r(p_{i})$ is a metric to reflect the impact of trivial words in semantic similarity between the sentence pair $p_{i}$, since higher $r(p_{i})$ indicates more trivial words are important towards semantic similarity calculation.
We randomly sample 300 sentence pairs from STS-B \cite{cer2017semeval} and select BERT as the USE, and we calculate the average \textbf{redundancy overlap ratio} $\hat{r}=\frac{\sum_{i=1}^{N} r(p_{i})}{N}$ w/o SR.
The results show that $\hat{r}$ reaches $10.2\%$ without SR, after applying SR, $\hat{r}$ drops to $7.1\%$\footnote{$\hat{r}$ changes since the inputs during similarity calculation have changed when SR activates. After SR. Eq~\ref{eq:h} becomes $H(x_{i}, x^{-}_{i}; w) = Sim(G(x_{i}),G(x^{-}_{i}))-Sim(G(x_{i}/w),G(x^{-}_{i}))$ where $G(\cdot)$ represents SR operation.} The results demonstrate that SR diminishes the impact of trivial words when measuring semantic similarity.

Moreover, we select some representative words and evaluate their importance w/o RepAL. As shown in Table~\ref{table:imp}, the results show that our SR indeed diminishes the impact of such trivial words when calculating semantic similarity.

\begin{table}[!h]\small
\centering
\begin{tabular}{@{}c|cc|c@{}}
\toprule
Word & No Refinement & With Refinement & $\Delta$ \\ \midrule
the  & 1.02        & 0.56     & -0.46    \\
a    & 0.98        & 0.43     & -0.55    \\
to   & 0.59        & 0.32     & -0.27    \\
in   & 0.68        & 0.21     & -0.47    \\
some & 0.60        & 0.31     & -0.29    \\
with & 0.72        & 0.24     & -0.48    \\
and  & 0.99        & 0.61     & -0.38    \\ \bottomrule
\end{tabular}
\caption{The importance of trivial words w/o sentence-level refinement. $\Delta$ means the importance change.}
\label{table:imp}
\end{table}

\subsection{Corpus-level Refinement}\label{corpuslevel}
To investigate whether corpus-level refinement diminishes the upper bound of the eigenvalue of embedding $E^{*}$, we make numerical experiments to dive into the relationship between the performance (Spearman correlation), $\lambda$ and the upper bound of the largest eigenvalue of $E^{*}$.

Specifically, we launch the experiments on six English benchmarks with BERT$_{base}$. As shown in Figure~\ref{figure:lamda}, when performance rises at peak, the upper bound of the largest eigenvalue of the embedding matrix $E^{*}$ is around the minimum, showing a coincidence between the two. The numerical results show that the corpus-level refinement enhances sentence embedding since it diminishes the largest eigenvalue of $E^{*}$. Previous methods \cite{huang2021whiteningbert} is equivalent to subtracting the average vector with $\lambda=1.0$, which fails to suppress the largest eigenvalue of embedding matrix extremely. However, our method chooses to subtract a larger $\lambda$ with adaptive weight, further suppressing the upper bound of the largest eigenvalue of the embedding matrix. The results show that the average embedding subtraction needs an adaptive weight. Moreover, this also illustrates why our method can still significantly improve the performance on W-BERT with substantial progress.
\end{document}